\documentclass{article}
\usepackage{graphicx}

\usepackage{tikz}
\usepackage{comment} 
\usepackage{amsmath,amssymb} 
\usepackage{color}

\newcommand{\vz}{{\bf z}}

\newcommand{\vx}{{\bf x}}
\newcommand{\vy}{{\bf y}}

\newcommand{\be}{\begin{equation}}
\newcommand{\ee}{\end{equation}}

\DeclareMathOperator*{\argmin}{arg\,min}

\begin{document}

\title{You Are Here:  Geolocation by Embedding\\ Maps and Images} 

\author{Noe Samano, Mengjie Zhou and Andrew Calway\\[2ex]
University of Bristol, UK}

\maketitle

\begin{abstract}
  We present a novel approach to geolocalising panoramic images on a 2-D cartographic map based on learning a low dimensional embedded space, which allows a comparison between an image captured at a location and local neighbourhoods of the map. The representation is not sufficiently discriminatory to allow localisation from a single image, but when concatenated along a route, localisation converges quickly, with over 90\% accuracy being achieved for routes of around 200m in length when using Google Street View and Open Street Map data. The method generalises a previous fixed semantic feature based approach and achieves significantly higher localisation accuracy and faster convergence.
\end{abstract}

\section{Introduction}
\begin{figure}[t]
\centerline{
\begin{tabular}{ccc}
\includegraphics[width=0.5\textwidth]{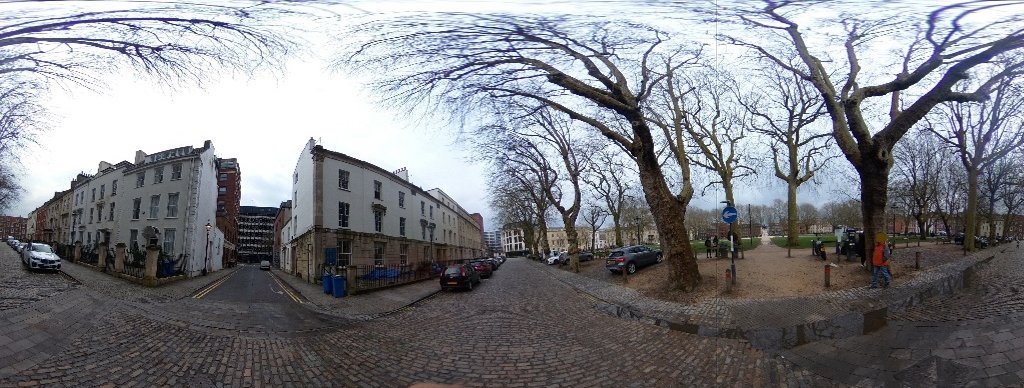} &
\includegraphics[width=0.2\textwidth]{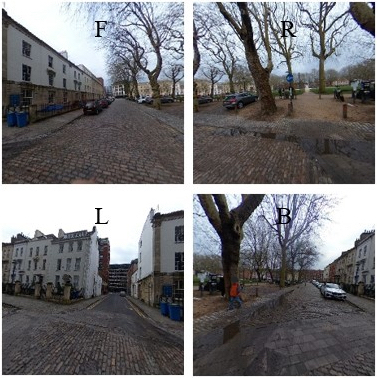} &
\includegraphics[width=0.2\textwidth]{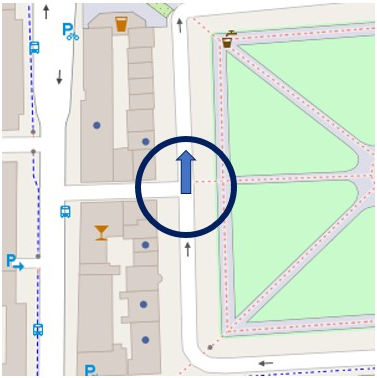} 
\end{tabular}
}
\caption{Images (centre) in different viewing directions within a panorama (left) captured at the location circled on the map (right), where the arrow indicates the heading direction (top left centre image). The junction, buildings and park area on the map are  visible in the images, illustrating the features we seek to leverage for geolocalisation.}
\label{fig:problem} 
\end{figure}

We consider the problem of geolocalising ground panoramic images on a 2-D cartographic map without using GPS, i.e. determining the geographic coordinates at which the images where captured. As illustrated in Figure \ref{fig:problem}, we seek to do this by linking the semantic information on the map to the content in the image, hence localising the latter. This is akin to the human skill of interpreting maps for way-finding using detailed survey maps or the ubiquitous You Are Here schematic maps found in cities and tourist attractions. It contrasts with previous works on image geolocalisation and the related problem of visual place recognition, which are based on matching the location image with a large database of georeferenced images, either ground or aerial (see Section \ref{sec:related}).

The motivation for using maps as the reference source, in preference to georeferenced images, is multifold. Comprehensive map data covering large areas is readily available, contrasting with the difficulty in obtaining sufficient ground images. Although large amounts of aerial imagery is available (in fact, it is often used to generate maps), using it or ground images for localisation means having to overcome the invariance challenges faced when matching images taken at the same location at different times. In contrast, maps provide a semantic representation which is to a large extent independent of capture time, and hence offers the potential for more robust matching. Techniques for relating image data to spatial semantics are also likely to be needed as human-robot interaction becomes increasingly sophisticated, such as endowing robots with the ability to produce visual descriptions to aid human way-finding.

We adopt a learning approach to the problem. It builds upon and generalises the work of Panphattarasap and Calway \cite{Panphattarasap2018}, who represent locations by binary descriptors indicating the presence or not of pre-selected semantic features (junctions and building gaps). This limits applicability to areas rich in those features and for which they provide sufficient discrimination. In contrast, we seek to learn descriptions that optimally link images to local map areas, without pre-assumptions as to which features are important, with the dual aim of generalising and increasing discrimination.

\begin{figure}[t]
\centerline{
\begin{tabular}{ccc}
\includegraphics[width=0.2\textwidth]{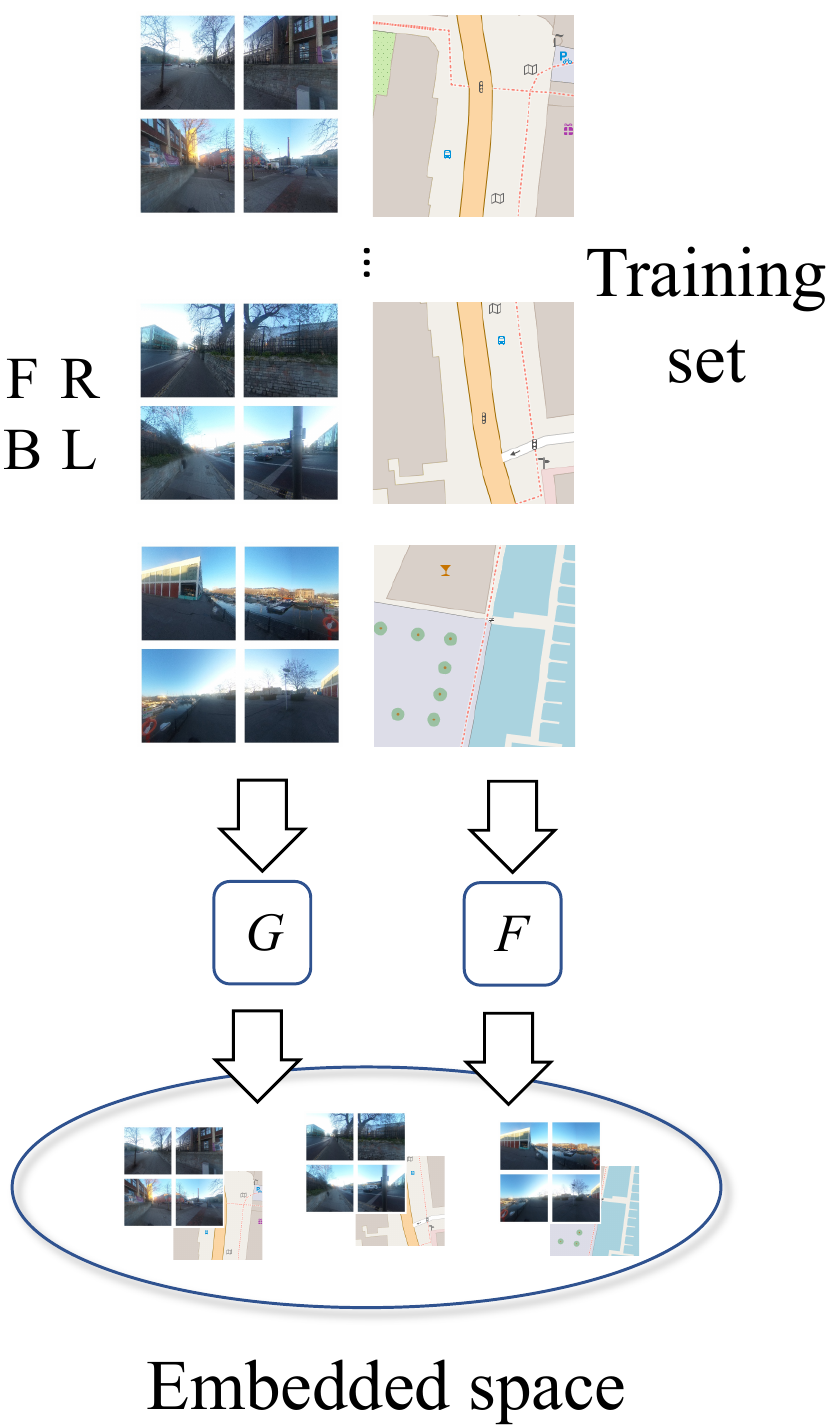} &&
\includegraphics[width=0.7\textwidth]{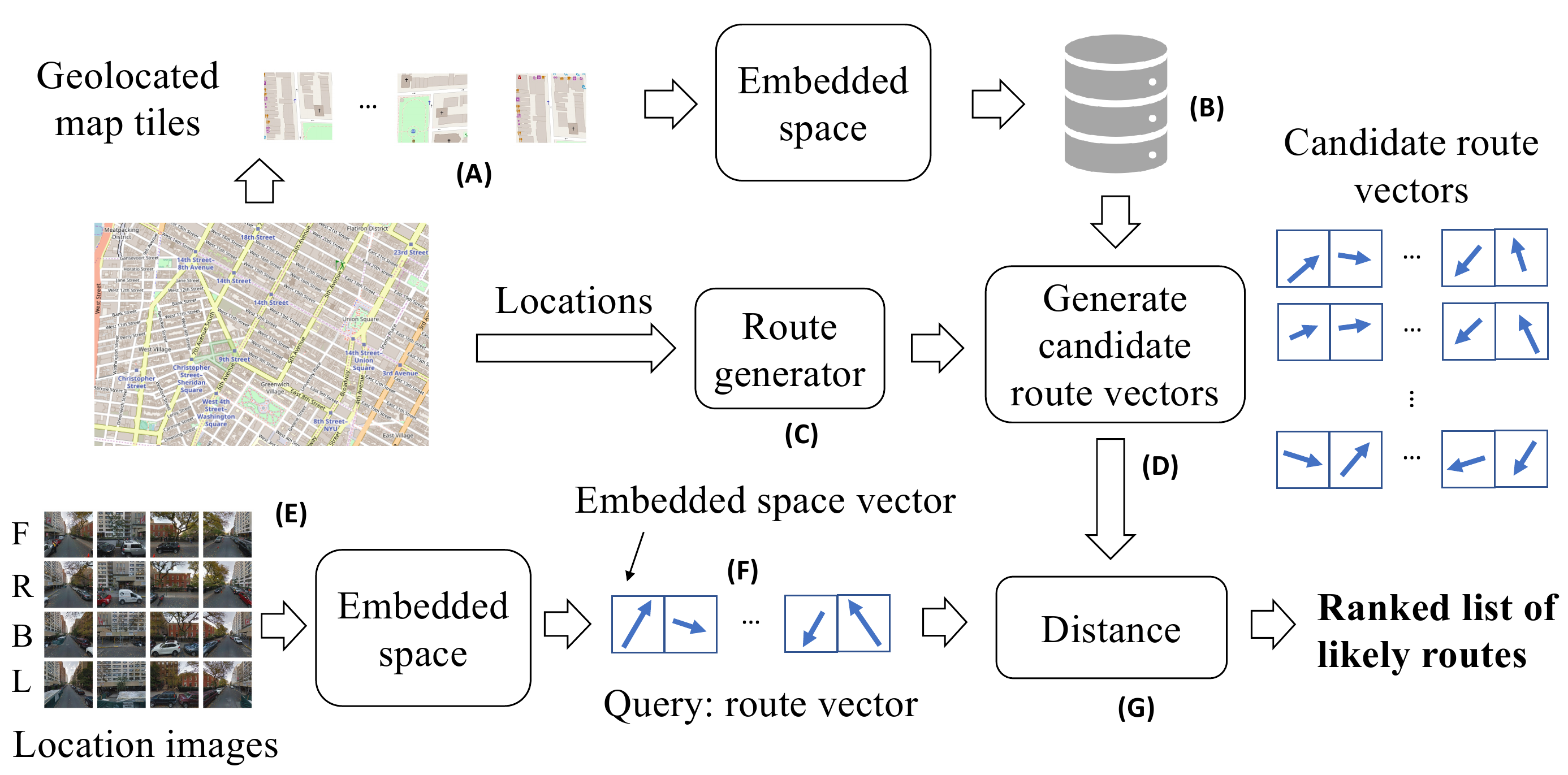}\\
(a) && (b)
\end{tabular}
}
\caption{(a) We learn transformations $G$ and $F$ which embed location images (in four orthogonal directions within a panorama - F, R, B and L) and map tiles into a low dimensional vector space (16-D), within which corresponding image/map tile pairs are close; (b) Geolocalisation: embedded vectors are computed for all map tiles (A), giving descriptors that are stored with their locations (B); all possible routes in the map are computed (C) and route descriptors generated from the map tiles (D); descriptors are computed for images along test routes (E) giving a query route descriptor (F), which is compared with all route descriptors (G), yielding a ranked list of likely routes.}
\label{fig:overview} 
\end{figure}

To do so, we learn a low dimensional embedded vector space (16-D) within which corresponding image and map tile pairs are close (a map tile is an image of a small region of the map), as illustrated in Figure \ref{fig:overview}a. Images from Google Street View (GSV)  \cite{streetlearnwebpage,mirowski2019streetlearn} and map tiles from Open Street Map (OSM) \cite{OSMwebpage} are used for training, and we use a Siamese-like network with triplet loss function to derive the embedded space. The network gives embedded space vectors for images and map tiles, which we treat as descriptors, and provide a means of assessing the similarity between an image and potential map locations. However, not surprisingly, the descriptors are not sufficiently discriminatory to localise a single image, since many places share similar map tiles.

Instead, analogous to that found in \cite{Panphattarasap2018}, it is the pattern of descriptors along routes that allows localisation, with the required route length dependent on local characteristics. Localisation therefore proceeds by comparing descriptors captured along a route with those derived from sequences of map tiles along all routes in the map, yielding a ranked list of likely locations and routes (see Figure \ref{fig:overview}b). As in \cite{Panphattarasap2018}, we consider the specific problem of geolocalising images known to lie at regular intervals along roads, mimicking images taken from a vehicle or by a pedestrian, for example, which allows us to carry out large scale experiments using GSV/OSM data. It also means that we can use a naive route matching algorithm to investigate the discriminatory power of the representation. In summary, we demonstrate the following key findings:
\begin{enumerate}
\item It is possible to learn an embedded space which links location images to map tiles, achieving top-1\% recall of between 72\%-82\%.
\item The space can be used to geolocalise image sequences along a route, achieving over 90\% accuracy for routes of approximately 200m within all testing areas of size 2-3 km$^2$. This compares with under 60\% when using the approach in \cite{Panphattarasap2018} for areas not rich in the chosen semantic features.
\item The approach gives increased accuracy for shorter routes and similar performance whether using turn information or not, demonstrating its ability to generalise and increase discrimination.
\end{enumerate}
We discuss related work next, followed in Section \ref{sec:learning} by details of the network architecture and learning. Section \ref{sec:localisation_process} gives details of the geolocalisation and Section \ref{sec:experiments} presents results of experiments using GSV and OSM data, including a comparison of performance with the method used in \cite{Panphattarasap2018}.

\begin{figure*}[t]
\centerline{
\includegraphics[width=0.8\linewidth]{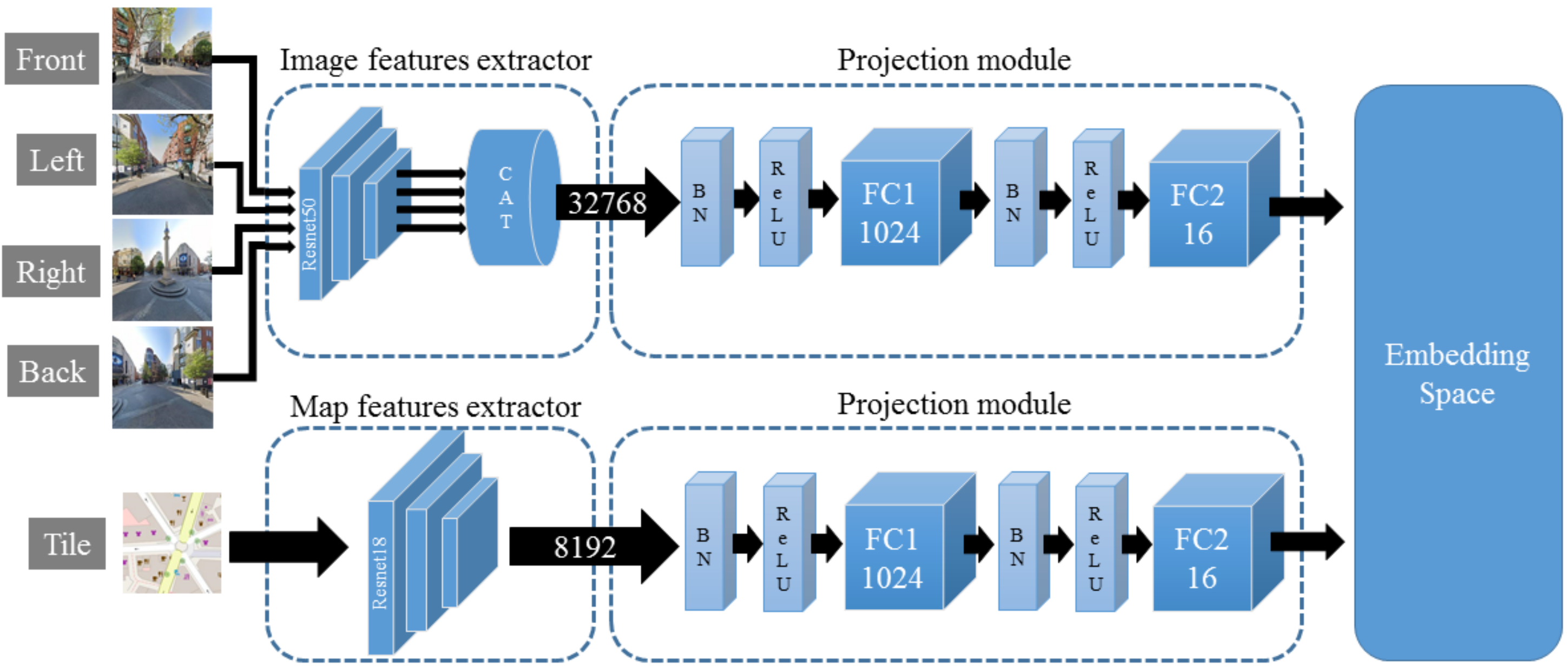}
}
   \caption{Network architecture for embedded space learning within which corresponding location 4-images (front, left, right and back viewing directions) and map tiles are close. Each is processed independently via a sub-network consisting of feature extraction and projection layers, resulting in a 16-D embedded space. See text for further details. }
\label{fig:net}
\end{figure*}

\section{Related Work\label{sec:related}}

The key insight identified in \cite{Panphattarasap2018} is that map semantics along a route uniquely identify location. 4-bit descriptors indicating the presence or not of semantic features (junctions and building gaps) are used to represent locations and convolutional neural network (CNN) classifiers trained on GSV/OSM data detect feature presence in orthogonal viewing directions. As in our approach, descriptors are concatenated along routes to give localisation by comparison with a database of route descriptors derived from the map. 85\% accuracy is reported for test routes up to 200m in length using GSV/OSM data when descriptors are combined with turn information along routes, although this was for a small test set (150 routes) within a densely built environment, which aligned well with the chosen semantic features. Notably, accuracy dropped to 45\% when turn information was not used. The same 4-bit descriptors are also used in a particle filtering implementation in \cite{Yan2019}, although results suggest that localisation is slower to converge, possibly due to the limited route memory within the filter. A key motivation of our work is to generalise these approaches to avoid the reliance on specific semantic features.

Semantic features are also used in \cite{castaldo2015} for geolocalising images by matching with GIS data. Coarse spatial layouts of known object classes are extracted from images and matched with GIS data at candidate locations using probabilistic descriptors. Top-20\% recall values of 70-80\% are reported, but this drops to $30\%$ for top-10\% recall, 
significantly lower than that achieved here and in \cite{Panphattarasap2018}, albeit for a much larger test area but smaller test set (50 locations).

Map data has been used for self-driving vehicle navigation \cite{Floros2013,Seff2016,Brubaker2016,Ma2017,Hecker2018,Amini2019,Ma2019}. In \cite{Seff2016}, learning from GSV/OSM data is used to recognise semantic features such as junctions, bike lanes, etc, to verify GPS, whilst in \cite{Floros2013,Brubaker2016} visual odometry is used to track map location by matching with road topology, extended in \cite{Ma2017} to incorporate semantic cues such as junctions and sun position. End-to-end learning is used in \cite{Hecker2018,Amini2019,Ma2019}, combining images with map road topology to predict steering control commands. Map semantics have also been used with GPS for 6-D pose estimation from images \cite{hentschel2010,Cham2010,Arth2015,Mousavian2016,Armagan2017a}, although the focus is on metric pose estimation as opposed to geolocalisation.

As noted above, our use of maps as a reference for geolocalisation contrasts with previous approaches, which are based on georeferenced ground or aerial images, see for example \cite{Hays2008,workman2015wide,Lin2015,vo2016localizing,Tian2017Cross-viewEnvironments,hu2018cvm,Hu2019} and the large body of work on visual place recognition \cite{Lowry-TOR-2015}. Many of these approaches learn embedded spaces to achieve ground-to-aerial cross-view matching, see e.g. \cite{Lin2015,workman2015wide,vo2016localizing,hu2018cvm,Hu2019}, and there is clear similarity with our approach since maps and aerial images are both overhead `plan views'. However, they differ in that maps provide a semantic representation as opposed to a time dependent snapshot of appearance, which as previously noted is an advantage when seeking to provide an invariant link with ground images. Nevertheless, given the similarities, we chose to adopt a similar network architecture and approach to learning, borrowing from that used in  \cite{Lin2015,workman2015wide,vo2016localizing,hu2018cvm,Hu2019}, but replacing aerial images with map tiles. We were also initially motivated by the work reported in \cite{Loc2vec}, which learns an embedded space for map tiles alone to reflect semantic similarity. 

\section{Embedding Maps and Images}
\label{sec:learning}
We now describe the model, data sets and methodology used to learn the embedded space. Our data inputs are panoramic images from GSV and RGB map tiles corresponding to local neighbourhoods of a 2-D map rendered from OSM.

\subsection{Network Model}

Our network architecture is shown in Figure~\ref{fig:net}. It has a Siamese-like form, with two independent sub-networks, one for location images and one for map tiles, each having a feature extractor and a projection module. We keep the weights in corresponding layers independent since each sub-network is processing information from very different domains.

In the location image sub-network, the feature extractor module is based on the Resnet50 architecture~\cite{he2016deep}, from which we have removed the average pooling and the last convolutional layers to produce a 4x4 feature map of 512-D local descriptors. The input to the module is four images corresponding to orthogonal front, left, right and back views (w.r.t the vehicle heading direction) cropped from a GSV panorama. Feature maps for each view are flattened and concatenated and then input to the projection module.

The map tile feature extractor also uses a residual network to produce a 512-D feature map, although we use Resnet18~\cite{he2016deep} in place of Resnet50, since map tiles contain considerably fewer details than location images. The module input is a map tile corresponding to the same location from which the GSV panorama was taken and the output is a descriptor vector. 

In both sub-networks, the projection module consists of two fully connected layers, both preceded by batch normalization~\cite{ioffe2015batch} and ReLu activation~\cite{nair2010rectified}, which reduce descriptor dimension down to the embedding size (we used 16-D) and help to project semantically similar inputs near to each other.

\subsection{Data sets}
\label{subsec:datasets}

For training and testing, we used the StreetLearn data set~\cite{streetlearnwebpage,mirowski2019streetlearn}, which contains 113,767 panoramic images extracted from GSV in the cities of New York (Manhattan) and Pittsburgh, U.S. Metadata is provided for all images, including geographic coordinates, neighbours and heading direction (yaw). Corresponding map tiles for each GSV image location were rendered from OSM using Mapnik \cite{Mapnikwebpage}, with all text and building numbering removed. 

To generate the training and testing datasets, we adopted the following procedure. First, using Breadth-First Search (BFS) and the same central panoramas as described in \cite{mirowski2019streetlearn}, we generated three testing subsets from areas in Hudson River (HR), Union Square (US) and Wall Street (WS), each containing 5,000 locations and covering $3.25\,km^2$, $2.77\,km^2$ and $2.33\,km^2$, respectively. The remaining locations in Manhattan, together with all locations from Pittsburgh, were used for training. For each location, we rendered two 256x256 pixel map tiles at different scales centred on the location coordinates and with the map projection aligned with the GSV heading direction to ensure a geographic match between the data domains. We used two scales to account for the fact that the relevance of semantic features in the map is dependent on visibility, i.e. a map tile at a larger scale may not include features visible in a location image, whereas a map tile at a smaller scale may include non-visible features. In this work we used map tiles with scales corresponding to geographic areas of approximately $152\times 152 \;m^2$ (small scale, S1) and $76\times 76\; m^2$ (large scale, S2). In total, the training  set consisted of 98,767 panorama images and 197,534 map tiles, two tiles for each location, and three testing data sets each with 5,000 panoramas and 10,000 map tiles.

\subsection{Training} 
\label{subsec:training}

We trained our model in an end-to-end way as all network parameters, including feature extraction, and projection layers, are updated at the same time. Since we do not have categories, or equivalently every location is a category, we train the model using an unsupervised method based on triplet loss metric learning~\cite{Schroff2015FaceNet}, similar to that used in \cite{hu2018cvm}. Note that we have to take care when considering data augmentation, since we need to maintain position and point of view relationships between the two domains, e.g. warping a map tile would require a suitable transformation of the corresponding location images, which is non-trivial to compute. Hence we limit augmentation to small changes in the scale of the map tiles and the viewing directions when cropping the panoramic images (see below). To form triplets, we take examples of matched and unmatched image/map tile pairs inside every batch.

We generate matched pairs as follows. Let $X$ and $Y$ denote the set of training tiles and panoramic images, respectively, and let $L$ be the set of all locations. We start the training process by taking $n$ random locations to form a subset $L_B$ of locations in the training batch. Then, for each location $l_i \in L_B$, we take the associated panoramic image $y_i \in Y$ and pick a map tile $x_i \in \{x^1_i,x^2_i\}$ at random, where $x^1_i$ and $x^2_i$ are the two map tiles at different scales for location $i$. We then apply data augmentation to generate $1\leq k\leq K$ different image/map tile pairs for each location $l_i$. Specifically, in the case of map tiles, we apply $T_x: x_i \to x_{ik}$, where $T_x$ applies a further random small scale change to $x_i$ and resizes it to 128x128 pixels, i.e. we randomly generate map tiles with random scales around that of the two `reference' tiles. In the case of the panoramic images, we apply $T_y : y_i \to y_{ik}=(y_{ik}^F,y_{ik}^L,y_{ik}^R,y_{ik}^B)$ that takes $y_i$ and crops four 128x128 pixel images in the front, left, right and back viewing directions, respectively, relative to the vehicle heading direction, with a random component in the cropping parameters to provide a degree of visual variation. We denote $y_{ik}$ as a `4-image'. In addition, we incorporate a standard colour normalisation and a random horizontal flip in both $T_x$ and $T_y$. 

In summary, at each training step, there are $N_B$ locations in the batch, each with $K$ 4-image/map tile pairs. This gives a total of $N_B^2K^2$ pairs, from which we can generate $N_BK^2$ matched pairs and $N_B(N_B-1)K^2$ unmatched pairs. Hence, the ratio of matched versus unmatched pairs depends on the number of locations in the batch. 

\subsection{Loss function}

For the loss function, we used the weighted soft-margin ranking loss proposed in~\cite{hu2018cvm}, a variation of that in~\cite{vo2016localizing,hermans2017defense} to address the problem of having to select the margin parameter in a traditional triplet loss~\cite{Schroff2015FaceNet}. It is defined as $\mathcal{L}_{wgt}(d) = ln(1+e^{\alpha d})$, where $d$ is the difference between a matched pair descriptor distance (Euclidean) and an unmatched pair descriptor distance, and $\alpha$ is a weighting factor that helps to improve convergence~\cite{hu2018cvm}. We also adopted a similar strategy to~\cite{Wang2016LearningEmbeddings} and included bidirectional cross-domain and intra-domain ranking constraints to force the network to preserve embedding structure in both data representations. Our loss function is therefore given by
\be
\begin{split}
\mathcal{L}(X,Y)=\sum_{i,j,k,l,m}\lambda_1 \mathcal{J}_{wgt}(\vx_{ik},\vy_{il},\vy_{jm})+\lambda_2 \mathcal{J}_{wgt}(\vy_{ik},\vx_{il},\vx_{jm})\\+
\sum_{i,j,k,l,m,k\neq l}\lambda_3 \mathcal{J}_{wgt}(\vx_{ik},\vx_{il},\vx_{jm})+\lambda_4 \mathcal{J}_{wgt}(\vy_{ik},\vy_{il},\vy_{jm})
\end{split}
\label{eq:softmargin_loss_with_constrains}
\ee
where $\mathcal{J}_{wgt}(\vx,\vy,\vz)=\mathcal{L}_{wgt}(d(\vx,\vy)-d(\vx,\vz))$, $\vx_{ik}$ and $\vy_{ik}$ denote the embedded vectors corresponding to the $k$th augmentation of the map tile and panoramic image at location $i$, respectively, i.e.\ the outputs from the sub-networks in Figure \ref{fig:net}, and $d(.,.)$ denotes Euclidean distance. Note that $i$ and $j$ refer to different locations, i.e.\ $1\leq i,j\leq N$ and $i\neq j$, and $k$, $l$ and $m$ refer to the $K$ augmentations per location, i.e.\ $1\leq k,l,m\leq K$. The values of $\lambda_1$, $\lambda_2$, $\lambda_3$ and $\lambda_4$ are weighting factors to control the influence of each constraint in the loss function. 

\subsection{Implementation}

To train our model, we initialized Resnet50 and Resnet18 with Places365~\cite{zhou2017places} and ImageNet~\cite{deng2009imagenet} weights, respectively. In the loss function, values of $\lambda_1 = \lambda_2 = \lambda_3 = \lambda_4 = 1.0$ achieved the best results. The embedding size was set to 16-D following ablation studies (see below) and we forced the embedded space to reside in an hyper-sphere manifold by performing L2-normalization on the network outputs and then scaling by a factor of 32. Although scaling does not affect nearest neighbor searching, it has an influence in the training process since it changes the steepness of the logistic curve \cite{vo2016localizing}. We set the value of $\alpha$ in $\mathcal{L}_{wgt}(d)$ to be 0.2, as we found that larger values in combination with a large scaling factor of the manifold, sometimes led to training collapse. The number of locations $N_B$ in a batch was set to 10, and the number of augmentations at each location $K$ was set to 5. Hence, in each learning step, there were 250 matched pairs and 2250 unmatched pairs. To form our triplets, we followed the batch all mining strategy discussed in~\cite{ding2015deep,hermans2017defense} and averaged the loss value over all possible triplets in the batch. We trained the model for 10 epochs using the Adam optimizer~\cite{kingma2014adam} with a $4\times10^{-5}$ learning rate.  

\section{Geolocalisation using embedded descriptors}
\label{sec:localisation_process}

We now describe using the learned embedded space for geolocalising images. The key principle is motivated by the approach used in \cite{Panphattarasap2018} and illustrated  in Figure \ref{fig:overview}b: along a route consisting of adjacent locations, the pattern of embedded descriptors obtained from the sequence of captured images enables the route to be uniquely distinguished from all other routes. The route length required for successful localisation is dependent on the semantic characteristics, which are reflected in both the images and the map. We emphasise again that we are considering the case of discrete locations at regular intervals along roads, enabling us to adopt the naive search approach to localisation described below.

Localisation proceeds as follows. Given a 4-image from each panorama along a route, we derive embedded descriptors via the learned model described above and compare the concatenation of the descriptors with those derived from sequences of map tiles corresponding to possible routes on the map. The closest gives the most likely route, as illustrated in Figure \ref{fig:overview}b. The methodology follows that in \cite{Panphattarasap2018} and so we provide a summary below and adopt similar notation for clarity.

For an area with $N$ locations, we define a route of length $m$ as a set of $m$ adjacent locations, $r_m=(l_{\gamma(1)},..,l_{\gamma(m)})$, where $l_{\gamma(i)}\in L$ for $1\leq \gamma(i)\leq N$ and the $\gamma(i)$, $1\leq i\leq m$, define the route trajectory. We limit the number of possible routes by considering only routes without loops or direction reversal. Given a panoramic image $y_{\gamma(i)}$ at location $l_{\gamma(i)}$, we extract a 4-image such that the front image (F) is aligned with the heading direction and use the learned model to obtain an embedded descriptor $\vy_{\gamma(i)}$ and hence a route descriptor $s^y_m=(\vy_{\gamma(1)},..,\vy_{\gamma(m)})$. Similarly, map routes with locations $\zeta(i)$ have route descriptors $s^x_m=(\vx_{\zeta(1)},..,\vx_{\zeta(m)})$ derived from the map tiles along the route. Note from earlier that these are also aligned with the heading direction. Let $S^x_m$ denote the set of route descriptors corresponding to all possible routes. Localisation then follows by finding the route descriptor $\hat{s}^x_m$ that is closest to $s^y_m$, i.e.\
\be
\hat{s}^x_m=\argmin_{s^x_m\in {\mathcal{S}^x_{m}}} DIST(s^x_m,s^y_m)
\label{eqn:closestroute}
\ee 
where $DIST(s^x_m,s^y_m)$ denotes the sum of the Euclidean distances between corresponding descriptors
\be
DIST(s^x_m,s^y_m)=\sum_{i=1}^m d(\vx_{\zeta(i)},\vy_{\gamma(i)})
\ee
In practice, we apply the above minimisation each time a new panoramic image is obtained along the route being traversed, giving an estimated route at each time step. This raises the question as to when successful localisation has been achieved, i.e.\ at what route length can we have sufficient confidence in the location estimate, which we investigate in the experiments. The results demonstrate that the required length of route depends on a number of factors, including area size, local appearance and environmental characteristics.

We also investigated the use of turn information along a route as a means of improving localisation accuracy, as adopted in 
\cite{Panphattarasap2018}. Specifically, again using similar notation to that in \cite{Panphattarasap2018}, we define a binary turn pattern along a route $r_m$ as $t_{m-1}=(t_{\gamma(1)},..,t_{\gamma(m-1)})$, where the $i$th component of $t$ indicates whether there is a turn between locations $l_{\gamma(i)}$ and $l_{\gamma(i+1)}$. Turn patterns can be estimated from the heading directions along a route and computed from OSM map data. This provides an additional constraint to that in Equation \ref{eqn:closestroute}, in that we require the best route estimate to have a matching turn pattern to that being traversed.

Finally, note that although we adopt a brute force search approach to the above, the recursive link between routes as the length increases means that it can be efficiently implemented by careful storage and searching of route indices $\zeta(i)$. There is also significant potential for culling routes that have low matching scores at each time step. We did a preliminary test of this idea in Figure \ref{fig:ESvsBSDdiff}b and intend to investigate it further in future work.

\section{Experiments}
\label{sec:experiments}

We evaluated the approach both in terms of the quality of the learned embedded space, i.e.\ its ability to link location images with appropriate map tiles, and the accuracy of localisation that can be achieved using embedded descriptors. We compared the latter with that achieved using binary semantic descriptors in \cite{Panphattarasap2018}. 

\subsection{Embedding maps and images}

\begin{figure}[t]
\centerline{
\begin{tabular}{ccc}
\includegraphics[width=0.45\textwidth]{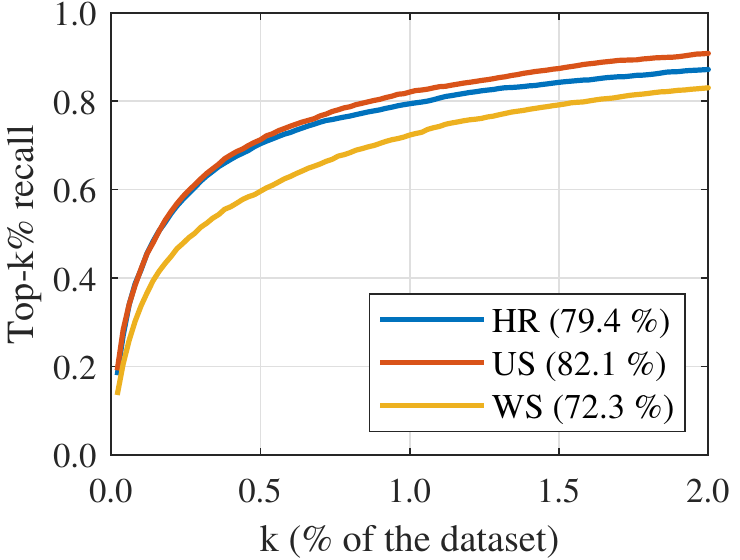} &&
\includegraphics[width=0.45\textwidth]{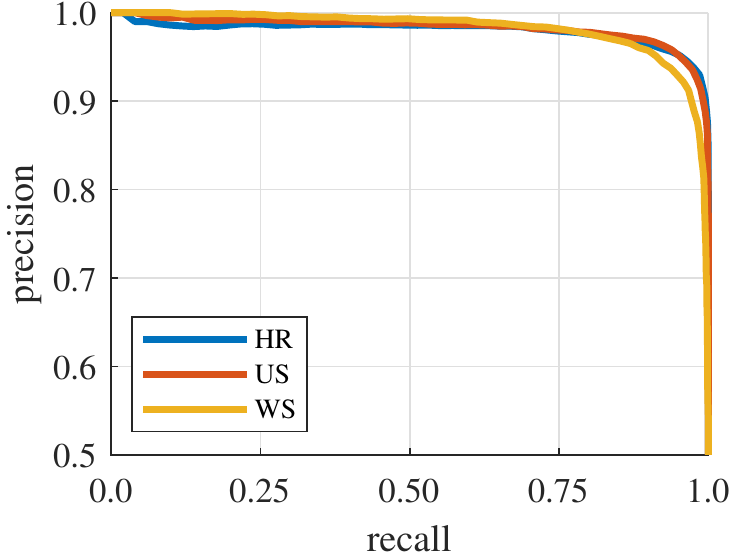}\\
(a) && (b)\\
\includegraphics[width=0.45\textwidth]{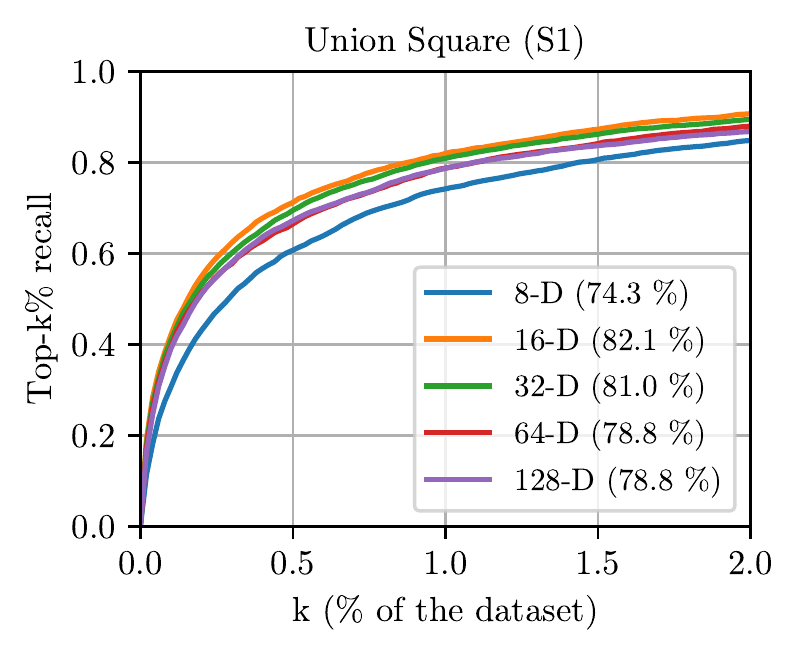} &&
\includegraphics[width=0.45\textwidth]{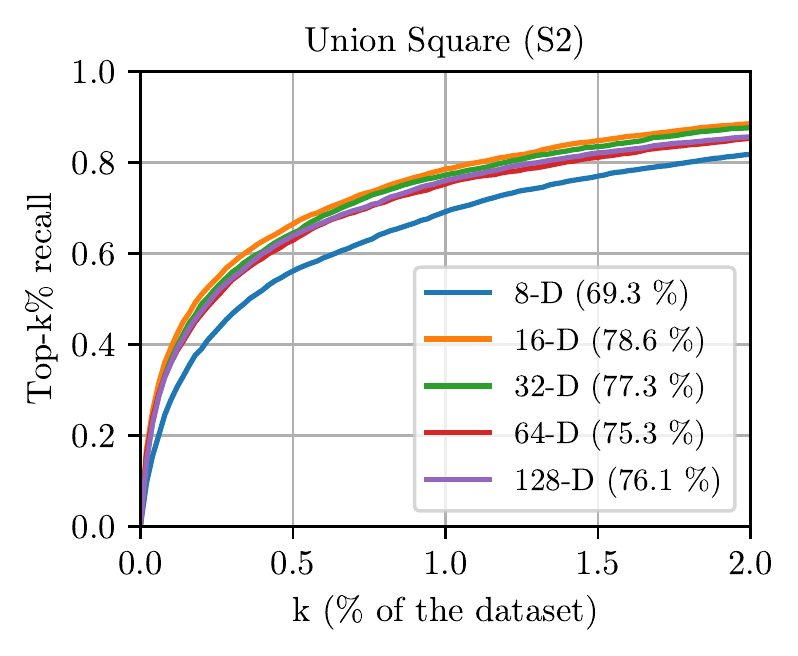}\\
(c) && (d)
\end{tabular}
}
\caption{a) Top $k$\% recall when learned embedded space is used to retrieve map tiles given a location image, where $k$\% is the fraction of the dataset size. Top 1\% recall values are shown in brackets; (b) Precision-recall curves generated by varying the distance cut-off threshold; (c-d) Top-k\% recall for Union Square using different space dimension sizes and map tile scales, S1 (c) and S2 (d).}
\label{fig:topk} 
\end{figure}

We investigated the recall performance when using the embedded space to retrieve corresponding map tiles given location images, i.e.\ how likely is the corresponding map tile to be the closest within the space. Top-$k$\% recall plots are shown in Figure~\ref{fig:topk}a, with top 1\% recall values in brackets, where top-$k$\% recall is the fraction of cases in which the ground truth tile is within the top $k$\% of best estimates. These results indicate that the network is doing well, with over 70\% top-1\% recall. This is confirmed in Figure~\ref{fig:topk}b, which shows precision-recall curves generated from balanced datasets of matched and unmatched pairs and by varying the distance cut-off threshold used when searching for close map tiles.

Note that performance is slightly worse for Wall Street, which reflects the fact that the area is noticeably different from those in the training set, with irregular intersections, motorways, and tunnels, making the area more challenging. We also noted significant mismatch between information in the map tiles and that in the location images, possibly resulting from redevelopment in the area yet to be reflected in OSM. Nevertheless, even under these  conditions, our model is able to assign high rank to corresponding map tiles.

\begin{figure}[t]
\centerline{
\begin{tabular}{ccc}
\includegraphics[width=0.45\textwidth]{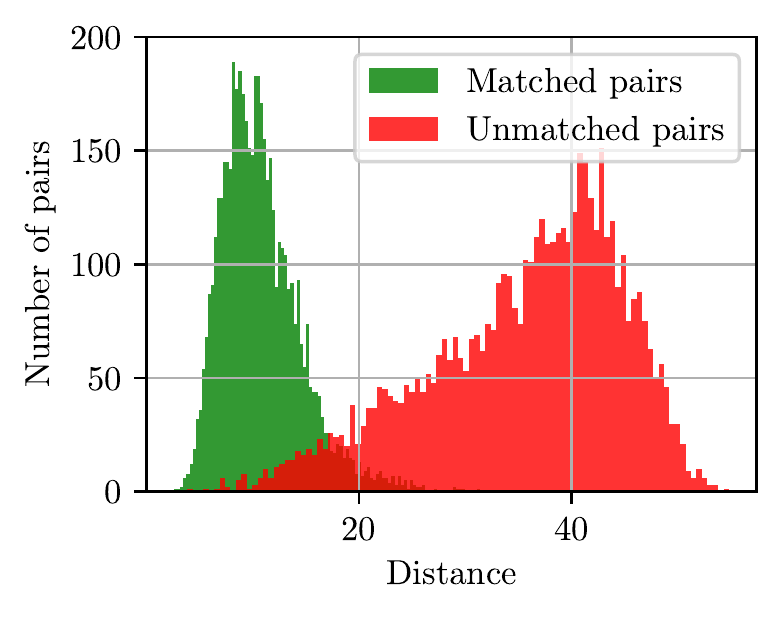} && 
\includegraphics[width=0.45\textwidth]{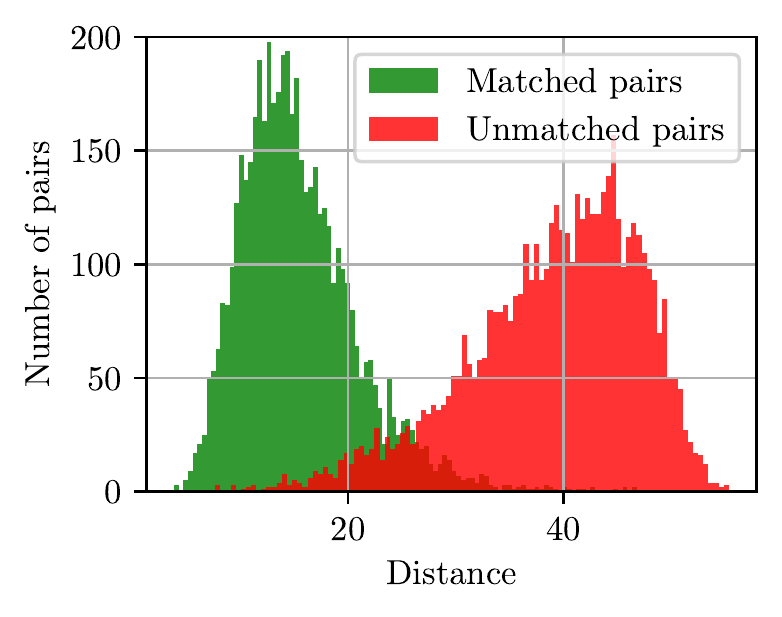} \\
(a) && (b)
\end{tabular}
}
\caption{Histograms of distance between matched (green) and unmatched pairs (red) in the embedding space for Union Square (a) and Wall Street (b). }
\label{fig:dist-hist} 
\end{figure}

\begin{figure*}[h!]
\begin{center}
\includegraphics[width=0.95\linewidth]{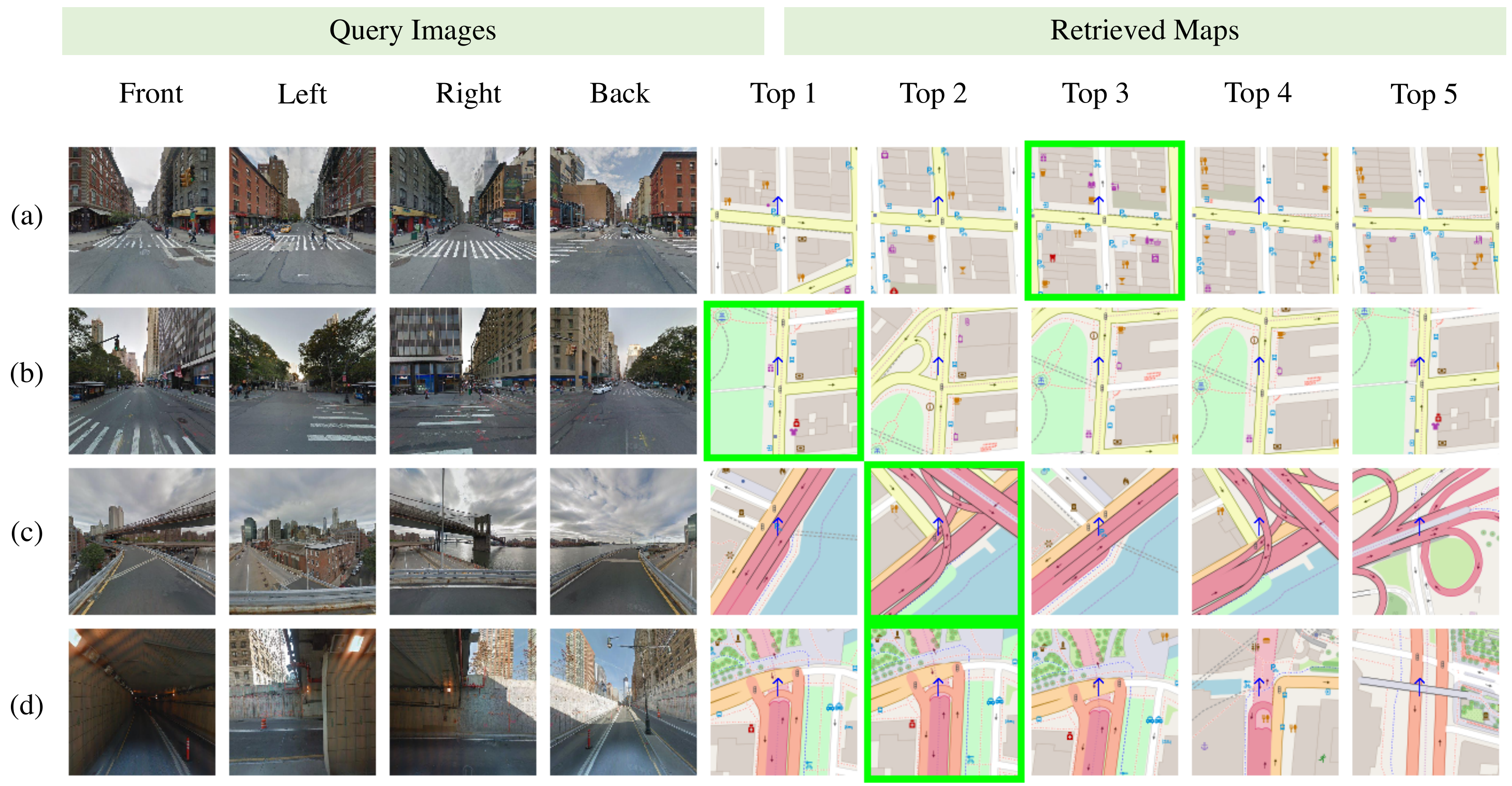}
\end{center}
   \caption{Top-5 retrieved examples (right) given a query 4-image (left) for Wall Street. The green frame encloses the true map location and the blue arrow in the center of the map tile represent the position and heading.}
\label{fig:retrieving_examples}
\end{figure*}

The above results were obtained using an embedded space dimension of 16 and using small scale (S1) map tiles when generating descriptors for localisation. We found that these values gave the best performance as illustrated in Figures \ref{fig:topk}c-d, which show the top-k recall when using different dimension sizes and when using the two scales for Union Square. Similar results were obtained for the other datasets and these settings were used in all of the  experiments.

We also examined how well the model separated matched and unmatched image/map tile pairs. Figure \ref{fig:dist-hist} shows histograms of the distance between matched and unmatched pairs inside the embedded space for Union Square and Wall Street. As expected, distances between unmatched pairs are larger, confirming that training has been effective and able to generalise outside the training set. Note also the larger overlapping area for Wall street, confirming the above comments regarding discrimination in the area.

To further illustrate the performance of the model, Figure \ref{fig:retrieving_examples} shows examples of query images and the top-5 retrieved map tiles for Wall Street, where the correct map tile is outlined in green. Note that the model has learned to relate semantics of the two domains, e.g.\ buildings (a-b), parks  (b), junctions (a-b), rivers (c), and tunnels  (d). In future we intend to investigate performance further, especially w.r.t optimising the architecture using appropriate ablation studies.

\subsection{Localisation}
\label{sec:localisation}

To evaluate geolocalisation performance, we randomly simulated 500 routes in each of the three test areas and applied the algorithm described in Section~\ref{sec:localisation_process}. In the route generation process, we excluded locations that were labelled as tunnels or motorways in OSM since we found that otherwise it generated many routes with very little appearance variation, consisting of long stretches tunnels or overpasses, which were near-impossible to distinguish from image data alone. To calculate localisation accuracy, we recorded the number of routes successfully localised at every step as a function of route length. We deemed a route to have been successfully localised if and only if the last 5 locations corresponds with those of the ground truth, i.e.\ the final 5 locations along the route. 

\begin{figure}[t]
\centerline{
\begin{tabular}{ccc}
\includegraphics[width=0.47\textwidth]{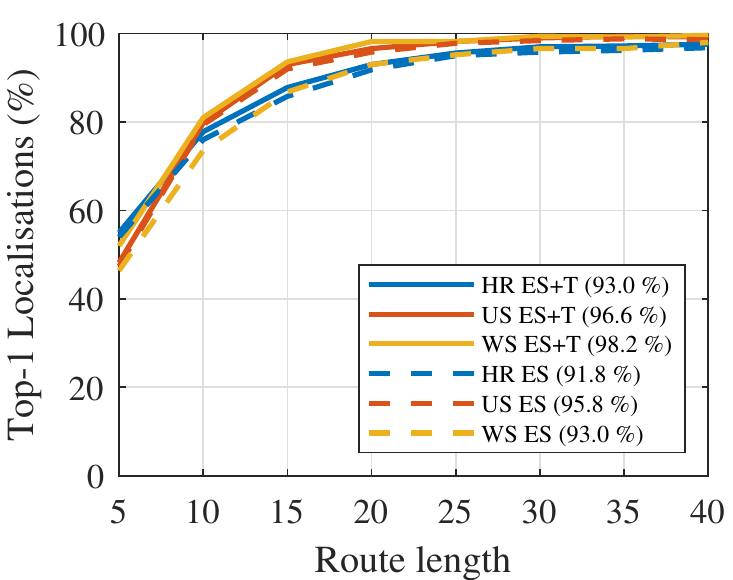} &&
\includegraphics[width=0.47\textwidth]{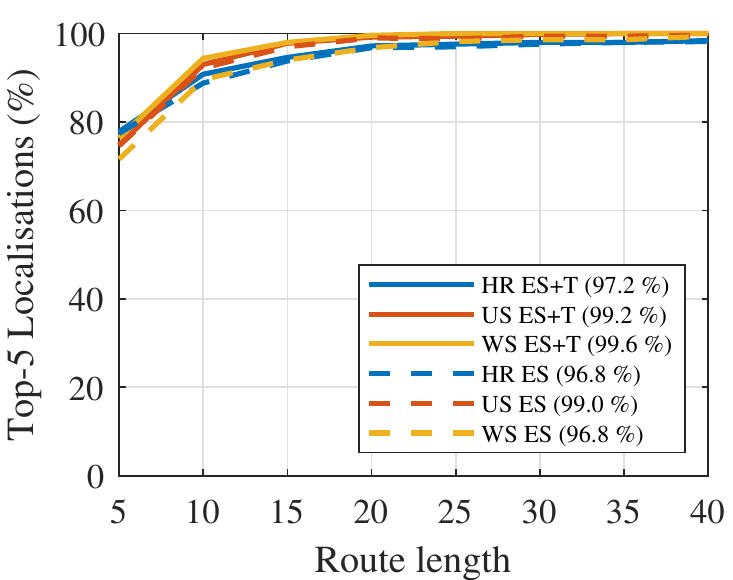}\\
(a) && (b)
\end{tabular}
}
\caption{(a) Top-1 and (b) top-5 localisation accuracy versus route length for all three datsets. Dashed lines indicate using embedded space descriptors only (ES) and solid lines indicate  including turn information (ES+T).}
\label{fig:ESres} 
\end{figure}

\begin{figure}[t]
\centerline{
\begin{tabular}{ccc}
\includegraphics[width=0.45\textwidth]{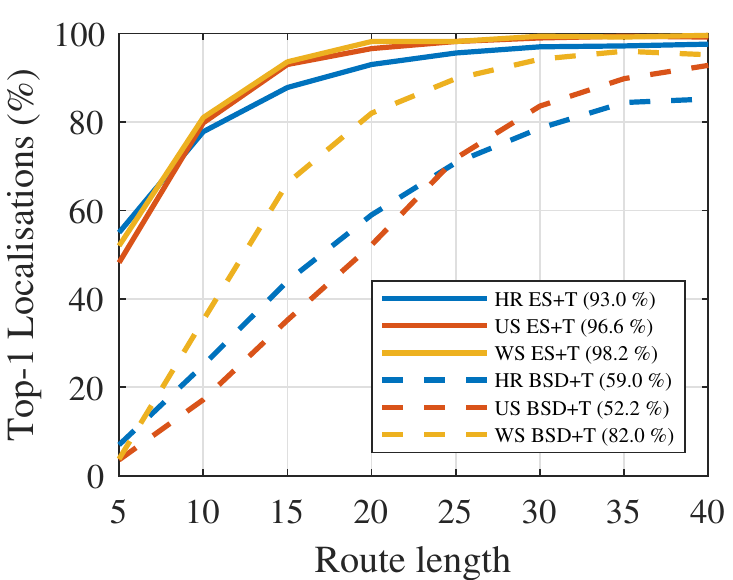} &&
\includegraphics[width=5.5cm]{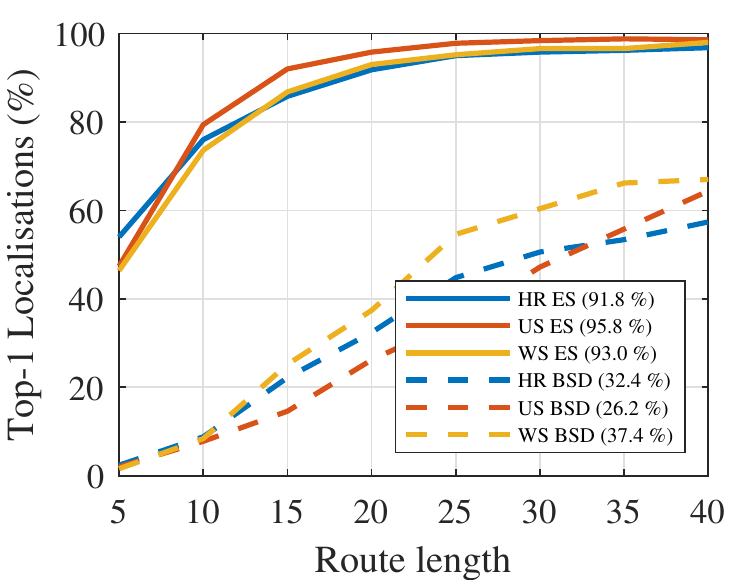}\\
(a) && (b)
\end{tabular}
}
\caption{Comparison of top-1 localisation accuracy between using our method and the BSD method in \cite{Panphattarasap2018} when a) using descriptors combined with turn information and (b) descriptors only.}
\label{fig:ESvsBSDturns} 
\end{figure}

Figure \ref{fig:ESres} shows top-1 and top-5  localisation accuracy versus route length when using embedded descriptors only and when combined with turn patterns. Top-$k$ accuracy for route length $m$ indicates the number of cases when a correct route, i.e. one meeting the above success criterion w.r.t the ground truth, was within the top $k$ most likely routes. Accuracy values for route lengths of 20 locations (approximately 200m) are shown in brackets. These results demonstrate that the method is performing exceptionally well, achieving over 90\% top-1 accuracy across all datasets for route lengths of 20, even when not using turn information. 

For shorter routes, although the top-1 accuracy drops, note from Figure \ref{fig:ESres}b, that the method still ranks correct routes highly, with over around 90\% within the top-5 for route lengths of 10. Note also that the overall performance of the method on the three test sets is comparable, indicating that the model is generalising well to unseen environments. It is also interesting to note that for Wall Street, the inclusion of turn information has a greater impact, reflecting the fact that the area contains significantly more variation in possible route configurations.

We also compared the method with the BSD approach in \cite{Panphattarasap2018}. We trained binary classifiers to detect junctions and building gaps, using the training set described in Section \ref{subsec:datasets} and labels obtained from OSM data in the same manner as described in \cite{Panphattarasap2018}. Experiments with different classifier architectures showed that Resnet18 \cite{he2016deep} gave the best performance, achieving for example precision/recall values of 0.6/0.75 and 0.74/0.8 for junctions and gaps, respectively, when tested on Hudson River, which are similar values to that given in \cite{Panphattarasap2018}. We used the classifiers to detect the presence or not of the features in each 4-image, i.e. junctions in the F and R images and gaps in the L and R images (cf. Section \ref{subsec:training}), and hence derived the 4-bit image-based BSD for each location. Map BSDs were generated in the same way as labels when generating the training set.

\begin{figure}[t]
\centerline{
\begin{tabular}{ccc}
\includegraphics[width=0.3\textwidth]{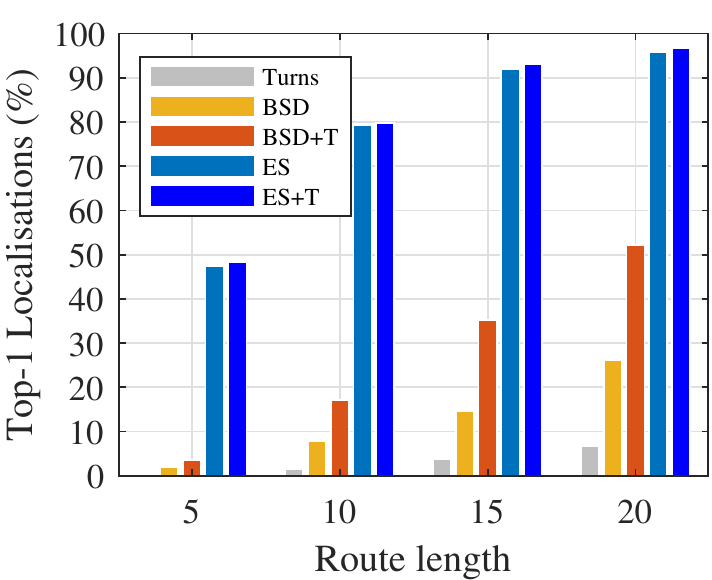} &
\includegraphics[width=0.3\textwidth]{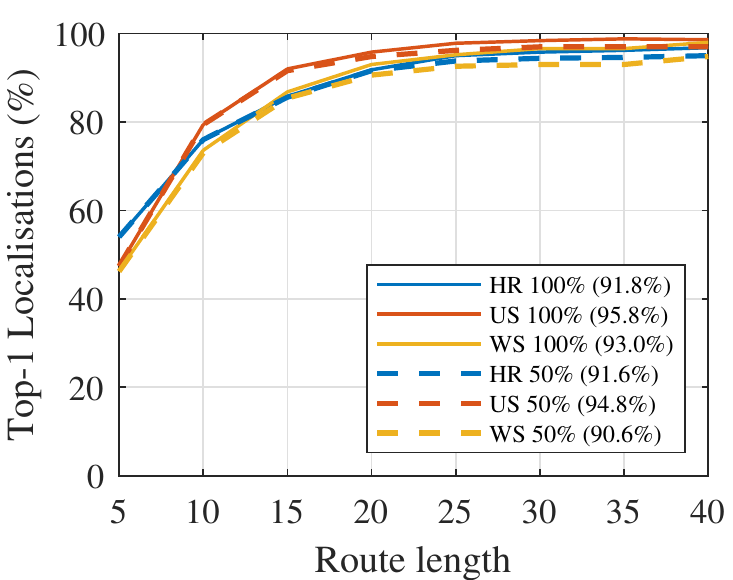} &
\includegraphics[width=0.3\textwidth]{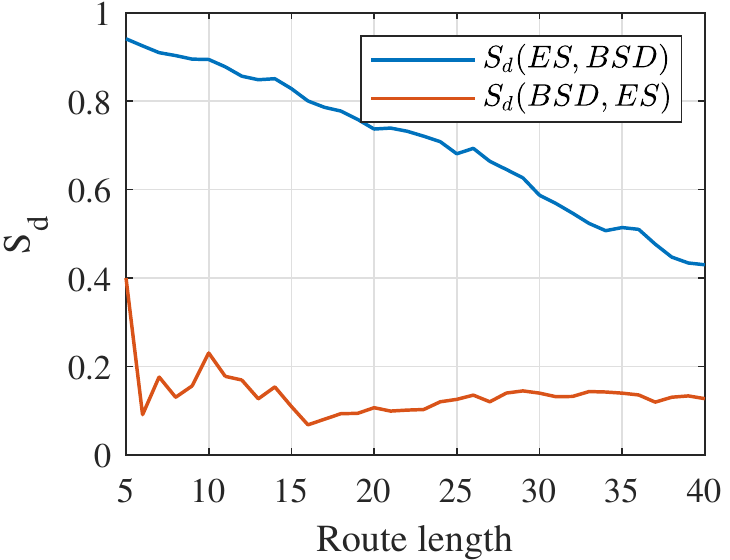}\\
(a) & (b) & (c)
\end{tabular}
}
\caption{(a) Top-1 localisation accuracy for 4 route lengths for the ES and BSD methods (with and without turn information) and when using only turn information for Union Square; (b) Accuracy when keeping 100\% (solid) and 50\% (dashed) of route candidates between updates; (c) Relative difference score between localisation sets for ES and BSD plotted against route length for Union Square. High scores for short routes suggests that our method is learning different semantic cues compared to BSD.}
\label{fig:ESvsBSDdiff}
\label{fig:onecol}
\end{figure}

Figure \ref{fig:ESvsBSDturns} shows top-1 accuracy versus route length using our method and the BSD method, both with and without turn information. Further comparison is shown in Figure \ref{fig:ESvsBSDdiff}a, which gives more detailed analysis of successful localisation rates for 4 route lengths. Also shown is the accuracy achieved when using only turn information, i.e.\ matching routes based only on the road pattern, which as also noted in \cite{Panphattarasap2018}, performs poorly, even for long routes.

These results demonstrate that our method gives superior performance when compared with using BSDs.  Although the latter performs reasonably well for Wall Street with turns, comparable to that reported in \cite{Panphattarasap2018} for a similar dense urban area with plenty of junctions and building gaps, its performance falls significantly in the other areas within which those features are less prevalent. In contrast, our method performs equally well in all areas, demonstrating that it is generalising well. Also notable is that the difference in performance is greater when not using turn information. We believe that this is due to the greater generality provided by the embedded space descriptors in contrast to the fixed features used in the BSD approach. Note also that our method is able to localise shorter routes with and without turns, again suggesting better generalisation.

We also investigated the robustness of our method by considering the impact of culling a significant percentage of route candidates in the searching process. Figure \ref{fig:ESvsBSDdiff}b shows a comparison of top-1 accuracy when using 100\% of route candidates and when discarding 50\% of them at each motion step until only 100 are left. Note the similarity of the curves, indicating that route discrimination occurs early and is maintained as routes grow.

Finally, to investigate the extent to which our model is using different information than BSD, we defined a score to measure the percentage of routes that our model is correctly localising but the BSD method is not and vice-versa. Formally, let $S_a$ and $S_b$ denote the set of routes successfully localised using methods $a$ and $b$, respectively, where $a,b \in$ \{BSD, ES\}. We define the difference score as $S_d(a,b) = | S_{a} \backslash S_{b} |/|S_{a}|$. A value of one would mean that none of the routes localised using method $a$ were localised by method $b$, whereas a value near zero would mean that all routes localised in $S_a$ were also in $S_b$. Results for both $S_d(ES,BSD)$ (blue) and $S_d(BSD,ES)$ (red) are shown in Figure \ref{fig:ESvsBSDdiff}c for Union Square. These results clearly show that our method is using different information, enabling it to generalise and increase discrimination. 

\section{Conclusions}

We have presented a novel method for correlating panorama location images and 2D cartographic map tiles into a common low dimensional space using a deep learning approach. This allows us to compare both domains directly using Euclidean distance. Furthermore, we have shown how this space can be used for geolocalisation using image to map tile matching defined along routes. Results indicate that the method can achieve over 90 \% accuracy when detecting routes of approximately 200 $m$ length in urban areas of over 2km$^2$.  Moreover, the approach significantly outperforms a previous method based on hand-crafted semantic features, demonstrating greater discrimination and generalisation.\\   

\noindent
\textbf{Acknowledgments.} We gratefully acknowledge the support of the Mexican Council of Science and Technology (CONACyT) and the Chinese Scholarship Council (CSC). We are also grateful to Google and the DeepMind StreetLearn project for use of the StreetLearn GSV dataset.

\bibliographystyle{abbrv}
\bibliography{references}
\end{document}